\documentclass[sigconf]{acmart}
\AtBeginDocument{%
  }

\copyrightyear{2026}
\acmYear{2026}
\setcopyright{cc}
\setcctype{by}
\acmConference[MM '26] {Proceedings of the 34th ACM International Conference on Multimedia}{November 10--14, 2026}{Rio de Janeiro, Brazil.}
\acmBooktitle{Proceedings of the 34th ACM International Conference on Multimedia (MM '26), November 10--14, 2026, Rio de Janeiro, Brazil}
\acmISBN{979-8-4007-2213-4/2026/11}
\acmDOI{10.1145/3767308.3837682}
\usepackage[capitalize]{cleveref}

\usepackage{multicol}
\usepackage{multirow}
\usepackage{xcolor}

\usepackage{bm}
\usepackage{graphicx}
\usepackage{stfloats}
\usepackage{algorithmic}
\usepackage{graphicx}
\usepackage{textcomp}
\usepackage{xcolor}
\usepackage{multirow}
\usepackage{makecell}
\usepackage{pifont}
\usepackage{xspace}
\usepackage{colortbl}

\usepackage{tabularx}
\usepackage{subcaption}
\usepackage{caption}
\usepackage{float}

\newcolumntype{Y}{>{\centering\arraybackslash}X}
\newcolumntype{P}[1]{>{\centering\arraybackslash}p{#1}}
\newcolumntype{C}[1]{>{\centering\let\newline\\\arraybackslash\hspace{0pt}}m{#1}}

\usepackage{pifont}%

\usepackage{booktabs}

\begin{document}

\title{Recognition-Conditioned Reasoning: A Training-Free \\ Multimodal-LLM Pipeline for Fine-Grained Micro-Action Understanding}

\author{Fengshun Wang}
\affiliation{%
  \institution{Wuhan University}
  \city{Wuhan}
  \country{China}
}
\email{ycwfs2001@163.com}

\author{Jin'ang Han}
\affiliation{%
  \institution{Wuhan University}
  \city{Wuhan}
  \country{China}
}
\email{pepperhan1@gmail.com}

\author{Zhigang Tu}
\authornote{Corresponding Author: tuzhigang@whu.edu.cn.}
\affiliation{%
  \institution{Wuhan University}
  \city{Wuhan}
  \country{China}
}
\email{tuzhigang@whu.edu.cn}


\begin{abstract}
Micro-actions are subtle, short, low-amplitude body movements, such as a fidgeting hand or a slight head tilt, that humans perform with little conscious intent yet that reliably leak emotional and psychological state. Understanding them goes beyond assigning a label: a model must also \emph{describe} which body parts move and \emph{reason}, faithfully, about why a clip warrants a particular fine-grained category. We present the training-free, prompt-only system that won first place in the fine-grained understanding track (MA-Bench) of the MAC~2026 Micro-Action Challenge, where both fine-tuning and ground-truth supervision are disallowed. Built entirely upon \emph{frozen} multimodal large language models (MLLMs), the system dynamically routes each of the eight sub-tasks to the MLLM empirically best suited for that task: a discriminative MLLM for closed-ended recognition tasks and a generative MLLM for open-ended description and reasoning tasks. This architecture achieves a statistically significant performance advantage on open-ended tasks, attaining an average score of 2.68 (on a five-point scale) compared to 1.44 for the second-best approach. 
To advance open-ended reasoning in multimodal learning, we propose \emph{recognition-conditioned reasoning}, a novel paradigm that decouples visual recognition from explanatory reasoning. Specifically, instead of relying on a single multimodal large language model (MLLM) to jointly perform clip labeling and free-text explanation, we leverage the coarse and fine-grained predictions from a discriminative MLLM to explicitly condition the reasoning process of a second, generative MLLM. To rigorously evaluate reasoning quality while avoiding the confounding biases introduced by LLM-based judges, we introduce a judge-free, gold-standard grounded evaluation metric that isolates reasoning fidelity from surface-level fluency. Our analysis reveals that residual reasoning errors are predominantly attributable to inaccuracies in discriminative labeling, not to perceptual limitations or linguistic fluency, and demonstrates that recognition conditioning improves the coarse and fine-grained label alignment of generated explanations. Critically, these gains are achieved without any fine-tuning, underscoring that principled, training-free orchestration of frozen MLLMs constitutes a robust, reproducible, and scalable pathway toward fine-grained micro-action understanding.
\end{abstract}




\maketitle

\section{Introduction}\label{sec:intro}

Humans constantly communicate through \emph{micro-actions} which include subtle, low-amplitude body movements like rubbing the hands, tilting the head, or shaking a leg, performed with little conscious intent yet reliably leaking emotional and psychological state~\cite{guo2024benchmarking}. Unlike the large, goal-directed activities of conventional action recognition~\cite{carreiraQuoVadisAction2017,feichtenhofer2019slowfast}, micro-actions are brief, fine-grained, and visually subtle, making them both a demanding test bed for video understanding and a valuable signal for affective computing, human-computer interaction, and mental-health screening~\cite{kolliasABAWValenceArousalEstimation2023}. The MA-52 benchmark~\cite{guo2024benchmarking} and the Micro-Action Analysis Grand Challenge series~\cite{guo2024mac,li2025mac} cast micro-actions as a concrete recognition problem. Its MAC~2026 ~\cite{li2026mac} successor, MA-Bench~\cite{li2026ma}, broadens the task from closed-set labelling toward open-ended, language-based \emph{understanding}.

\textbf{Micro-action understanding} demands far more than a label. Across $12{,}000$ video questions spanning eight sub-tasks, a model must recognise the coarse and fine category (multiple choice), answer fine-grained spatial-temporal judgements (yes/no), \emph{describe} which body parts move and how, and \emph{reason} about why a clip warrants a particular label. The open-ended answers are scored by language judges that rate spatial, temporal, and causal faithfulness. This mirrors a broad shift from specialised video classifiers toward multimodal large language models (MLLMs) as general-purpose video understanders~\cite{liMVBenchComprehensiveMultimodal2024a,xuSlowFastLLaVAStrongTrainingFree2024}. Crucially, the challenge forbids any fine-tuning or use of ground-truth labels: the models must be used \emph{as is}, steered only through prompting.

Frozen MLLMs are an awkward fit for this regime. They perceive coarse scene content but routinely confuse neighbouring fine categories (tilting versus turning the head, touching versus scratching) and, when asked to reason, emit fluent explanations that nonetheless assert the wrong label~\cite{jiangMMECoTBenchmarkingChainofThought}. Worse, the fluency-biased language judges that score the open-ended answers reward the well-written narrative and partly mask the label error~\cite{chenMLLMasaJudgeAssessingMultimodal2024}, so a system can look strong while being quietly wrong about the action itself. No single off-the-shelf model is uniformly best, either: the model that picks the correct multiple-choice label is rarely the one that writes the most faithful description.

We turn these obstacles into design principles. First, rather than committing to one MLLM, we \emph{route} each sub-task to the frozen model empirically strongest on it: a discriminative MLLM for the closed-ended recognition sub-tasks and a generative MLLM for the open-ended description and reasoning sub-tasks. Second, we observe that asking one MLLM to both \emph{discriminate} the label and \emph{explain} it conflates two competencies it possesses to very different degrees. A controlled, gold-grounded probe makes this precise: our generated explanations are spatially and temporally faithful, yet a sizeable fraction still assert a neighbouring fine category, so the residual bottleneck is discriminative label accuracy rather than perception or fluency. We therefore propose \textbf{recognition-conditioned reasoning}: inject the discriminative MLLM's predicted coarse and fine label into the generative MLLM's prompt to \emph{condition} its free-text explanation. To measure this without trusting a biased judge, we introduce a gold-grounded \emph{label-mention} metric that checks, against cross-task ground truth, whether a generated explanation actually asserts the correct category. Under this metric, conditioning lifts coarse and fine label accuracy by $23$ and $28$ points.

This training-free pipeline wins \textbf{first place} on MA-Bench at MAC~2026. Tellingly, the victory is created almost entirely on the open-ended description and reasoning sub-tasks (open-ended average $2.68$ versus $1.44$ for the runner-up), even though our closed-ended recognition slightly trails. Careful, training-free orchestration of frozen MLLMs, not bespoke training, is what delivers fine-grained micro-action understanding.

\noindent Our main contributions are summarised as follows.

\begin{itemize}
\item We present the first-place, training-free system for fine-grained micro-action understanding at MAC~2026: a prompt-only pipeline over frozen MLLMs that routes each of the eight sub-tasks to the model best suited to it.
\item We introduce \emph{recognition-conditioned reasoning}, which decouples discrimination from explanation by conditioning a generative MLLM's reasoning on a discriminative MLLM's predicted label, and we trace the micro-action reasoning bottleneck to label accuracy rather than perception or fluency.
\item We propose a judge-free, gold-grounded label-mention metric for open-ended micro-action reasoning that avoids LLM-judge fluency bias and validates a $+23/+28$ point gain in coarse and fine label accuracy.
\item We analyse the practical levers of a frozen-MLLM pipeline (per-task model choice, frame sampling, image fidelity, and prompt design) and report the configuration that won the challenge.
\end{itemize}
\vspace{0.8\baselineskip}

\section{Related Work}\label{ref:rel}

\paragraph{Micro-Action and Nonverbal Behavior Understanding.} Micro-actions sit between two well-studied extremes: the large, intentional activities of action recognition and the facial muscle movements of micro-expression analysis. They are whole-body but low-amplitude and largely involuntary, which is precisely what makes them informative about affect~\cite{kolliasABAWValenceArousalEstimation2023}. The MA-52 benchmark~\cite{guo2024benchmarking} established a 52-class taxonomy for single-label micro-action recognition, and MMA-52~\cite{li2025mmad} extended it to multi-label temporal detection. The MAC~2026 MA-Bench track~\cite{li2026ma} adds open-ended description and reasoning on top, continuing the Micro-Action Analysis Grand Challenge series~\cite{guo2024mac,li2025mac,li2026mac}. A growing body of dedicated micro-action and micro-gesture methods spans skeleton-based recognition~\cite{li2025prototypical,gu2025motion,chen2024prototype,li2023joint}, online recognition~\cite{liu2024micro,liu2025online,shen2026spatial}, and multimodal or self-supervised fusion~\cite{gu2025mm,liu2026self,shang2025cross,li2023data,guo2026rethinking,cheng2025owlsight}. Like Micro-DualNet~\cite{chappaMicroDualNetDualPathSpatioTemporal2026} and the challenge-winning data-centric systems~\cite{wangCombattingDataImbalance2025}, these train task-specific networks, whereas we keep the backbone frozen and adapt only a lightweight probe or the prompt. And there are a few winner methods ~\cite{wang2024instance,gong2024micro,yu2024end,wang2025combatting,li2025progressive,li2024advancing}. More broadly, recent work grounds nonverbal communication and social signals in MLLM reasoning~\cite{kimGRASPLearningGround2026,lanExpLLMChainThought2024}, motivating the language-based understanding that MA-Bench evaluates.

\paragraph{Video Foundation Models.}
Self-supervised pretraining has produced general video encoders that now serve as the visual backbones of multimodal LLMs, including masked autoencoding~\cite{tong2022videomae,wangVideoMAEV2Scaling2023}, multimodal contrastive scaling~\cite{wangInternVideo2ScalingFoundation2024a}, and joint-embedding predictive models such as V-JEPA2~\cite{assranVJEPA2SelfSupervised2025}. Because such encoders are costly to adapt and our setting forbids fine-tuning, we use the resulting models strictly \emph{frozen} and place all task-specific capacity in prompting. Lightweight-probe adaptation of frozen transformers~\cite{chenAdaptFormerAdaptingVision2022b} reflects the same training-light philosophy.

\paragraph{Multimodal LLMs for Video Understanding.}
MLLMs have rapidly become general video understanders, spanning open families such as Qwen~\cite{yangQwen2TechnicalReport2024}, InternVL~\cite{chenInternVLScalingVision,chenHowFarAre2024}, and compact models like MiniCPM-V~\cite{yaoMiniCPMVGPT4VLevel2024}. A particularly relevant line keeps the model entirely frozen and feeds sampled frames as a prompt, trading training for inference-time design~\cite{xuSlowFastLLaVAStrongTrainingFree2024,xuPLLaVAParameterfreeLLaVA2024}, and frozen generative priors have even been repurposed for other video tasks with no task supervision~\cite{zhang2026leveraging}. Frame-selection strategies then become a central lever~\cite{tangAdaptiveKeyframeSampling}. Benchmarks such as MVBench~\cite{liMVBenchComprehensiveMultimodal2024a} reveal that these models excel at coarse semantics but remain weak on fine-grained and temporally precise judgments, precisely the regime that micro-actions inhabit. Because the MA-Bench rules forbid fine-tuning, our system is purely prompt-based, and we exploit, rather than fight, the heterogeneous strengths of different frozen MLLMs.

\paragraph{Prompting and LLM-based Evaluation.}
Prompting and chain-of-thought elicitation improve LLM reasoning~\cite{diaoActivePromptingChainofThought2024}, and step-by-step schemes have been ported to video~\cite{feiVideoofThoughtStepbyStepVideo}, while inference-time visual prompting steers frozen generative video models without fine-tuning~\cite{zhang2025visual}. Yet multimodal chain-of-thought can also \emph{hurt} perception-heavy tasks~\cite{jiangMMECoTBenchmarkingChainofThought}, and explanation quality can diverge from label correctness, a separation made explicit in explainable action assessment~\cite{qiExplainableActionForm2025}. These observations motivate decoupling discrimination from explanation, which our recognition-conditioned reasoning realises as a training-free cascade between two frozen MLLMs, conditioning a generator on a discriminative prediction. 

\section{Methodology}
\label{sec:methodology}

\begin{figure*}[t]
\centering
\includegraphics[width=\textwidth]{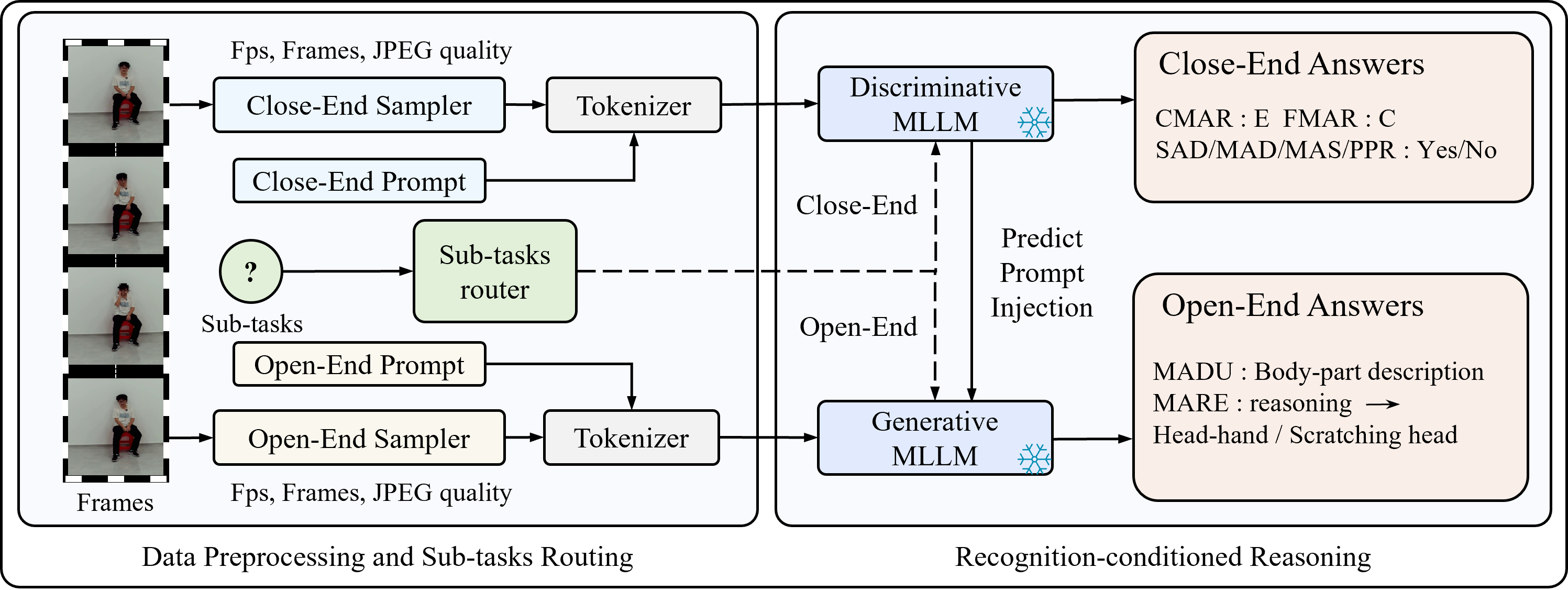}
\caption{Our training-free pipeline for micro-action understanding. Sampled frames and the prompts are routed by sub-task to the frozen MLLM best suited to it, a discriminative model (Gpt-5.5) answers the closed-ended recognition and judgement sub-tasks, and a generative model (Qwen3.7-plus) produces the open-ended description (MADU) and reasoning (MARE). In \emph{recognition-conditioned reasoning}, the discriminative model's predicted coarse and fine label is injected to condition the generative model's explanation.}
\label{fig:overview}
\end{figure*}

\subsection{Task Formulation and Overview}
\Cref{fig:overview} shows the pipeline. We keep every model entirely \emph{frozen} and place all task-specific capacity in prompt-level orchestration around it. MA-Bench pairs each short single-person clip with a question drawn from eight sub-tasks: two are multiple-choice recognition (coarse CMAR, fine FMAR), four are yes/no spatial-temporal judgements (SAD, MAD, MAS, PPR), and two are open-ended generation, namely describing which body parts move (MADU) and reasoning out the correct labels (MARE). Closed-ended sub-tasks are scored by accuracy. Each open-ended sub-task is scored by a language judge along three dimensions. For MADU these are behavioural-semantic, spatial-topological, and temporal-structural. For MARE they are coarse-label, fine-label, and causal-reasoning. The final score is the average of all twelve sub-scores after normalising each to $[0,100]$, with the open-ended scores in $[0,5]$ multiplied by $20$. Fine-tuning and any use of ground-truth labels are not allowed, so the system is built entirely by selecting and prompting frozen Multimodal LLMs.

\subsection{Training-Free Micro-Action Understanding}
For each question we uniformly sample frames from the clip, encode them as images, and pair them with a per-sub-task system prompt that fixes the output format (a single option letter, a \texttt{Yes}/\texttt{No} token, or free text). Two design choices matter most. \textbf{Per-sub-task model routing:} rather than committing to one MLLM, we benchmark several frozen models on a development split and assign each sub-task to the model that is empirically strongest on it, a discriminative MLLM for the multiple-choice and yes/no recognition sub-tasks, and a generative MLLM for the open-ended description and reasoning sub-tasks. Orthogonally, the temporal-order sub-tasks receive a larger frame budget, which short samples otherwise starve of ordering cues, as detailed in \cref{sec:frames}. Routing each input to the handler best suited to it parallels difficulty- and type-adaptive routing for zero-shot video understanding~\cite{zhang2026dart}. 

Our recipe is two-stage. \emph{Offline}, we select the strongest frozen model for each sub-task on the labelled development split, as reported in \cref{tab:models}. \emph{Online}, we refine the prompts against the public leaderboard. We sample frames at a target rate of $3$~fps, capped at $8$ frames per clip, raised to $64$ for the temporal-order sub-tasks MAS and MAD as \cref{sec:frames} explains, and JPEG-encode them at quality $95$. Decoding budgets are routed by answer type: $32$ tokens for closed-ended answers and $1024$ tokens for open-ended description and reasoning, which need room for a full explanation. The prompts evolved over four iterations, each targeting specific scoring dimensions summarised in \cref{tab:prompts}. v2 fixes the output format. v3 adds explicit body-part, direction, and trend cues. v4 rewrites the open-ended prompts around the official rubric, namely a $\geq 3$-evidence causal chain, near-neighbour exclusion, and an explicit final label line. v5 adds recognition-conditioning, described in \cref{sec:rcr}. 
Each open-ended prompt is written to cover exactly the three dimensions its language judge scores, and v5 appends an explicit label line to the MARE prompt to carry the injected coarse/fine prediction. \Cref{fig:prompts} shows these deployed prompts verbatim.

\begin{figure}[t]
\centering
\small
\setlength{\fboxsep}{5pt}
\fbox{\begin{minipage}{0.93\columnwidth}
\textbf{CMAR (closed)}: \emph{``\dots Answer with EXACTLY ONE capital letter from A to G. Output only that letter, with no explanation, punctuation, or extra words.''}
\end{minipage}}\\[3pt]
\fbox{\begin{minipage}{0.93\columnwidth}
\textbf{MADU (describe)}: \emph{``Describe the person's MOVEMENT in plain prose\dots judged on three things, so cover ALL: (1)~\textbf{Dominant action}: the dominant body part and the specific micro-action, the lead-follow relation, and whether the motion is single/continuous/repetitive; (2)~\textbf{Spatial detail}: direction, position relative to the torso/midline, and the motion PATH (start$\to$end); (3)~\textbf{Temporal structure}: phases, order, rhythm. 3--6 sentences.''}
\end{minipage}}\\[3pt]
\fbox{\begin{minipage}{0.93\columnwidth}
\textbf{MARE (reason)}: \emph{``\dots(1)~\textbf{Causal chain}: state at least THREE concrete observations (part; direction/trend; contact; temporal pattern), then RULE OUT the most likely look-alike, then conclude; (2)~\textbf{Coarse label}: the dominant body-part category, using the contact category (Head-hand, Leg-hand, Body-hand) when parts touch; (3)~\textbf{Fine label}: the exact micro-action with direction/trend and contact, distinguished from close neighbours (Tilting vs.\ Turning, Touching vs.\ Scratching).''} \;\textbf{+v5:} \emph{end with} \texttt{[coarse]\,<cat>\,|\,[fine]\,<action>} \emph{on the last line.}
\end{minipage}}
\caption{Representative deployed system prompts, abridged verbatim from our submission. The open-ended MADU and MARE prompts are each structured to cover the three dimensions their judge scores, and the \textbf{+v5} line shows the optional label appended for recognition-conditioning.}
\label{fig:prompts}
\end{figure}

\subsection{Recognition-Conditioned Reasoning}
\label{sec:rcr}
Asking a single MLLM to \emph{both} name a micro-action and \emph{explain} it conflates two competencies it possesses unequally, it narrates motion fluently but frequently commits to a wrong neighbouring fine label (tilting vs.\ turning, touching vs.\ scratching). Fluency-biased language judges reward the well-written narrative and partly mask the label error~\cite{chenMLLMasaJudgeAssessingMultimodal2024,jiangMMECoTBenchmarkingChainofThought}. We therefore \emph{decouple} the two. A discriminative MLLM first predicts the coarse and fine labels, drawing on its multiple-choice strength. The predicted label is then injected into the generative MLLM's reasoning prompt as a strong prior, which it must justify or, if the evidence contradicts it, override. This is a training-free cascade between two frozen models, requiring no ground truth at all.


To evaluate the generated reasoning without trusting a biased judge, we exploit a structural property of MA-Bench: because the benchmark reuses each video across sub-tasks, the gold coarse and fine label for a reasoning item is recoverable from the same video's recognition items. We parse the label asserted by the generated explanation and check it for exact (synonym-normalised) agreement with this cross-task gold, yielding a \emph{judge-free label-mention accuracy} for the two label dimensions of MARE. This metric isolates the label bottleneck and quantifies the effect of conditioning. \Cref{sec:rcr-study} reports the result.

\section{Experiments}\label{ref:exp}

\subsection{Dataset and Evaluation}
We evaluate on \textbf{MA-Bench}~\cite{li2026ma}, the MAC~2026 micro-action understanding benchmark: $12{,}000$ video questions over the eight sub-tasks of \cref{sec:methodology}, built on the $1{,}000$ short single-person clips of the MA-52 family~\cite{guo2024benchmarking}. Submissions are scored by the official six closed-ended accuracies and six open-ended language-judge dimensions, summarised by the final twelve-way average. We tune every design choice on the labelled development split, since the released test labels are placeholders.

\paragraph{Local evaluation protocol.}
We score the development split with the same recipe as the official metric, so that model and prompt selection track the leaderboard. Closed-ended sub-tasks use exact-match accuracy. Each open-ended answer is graded by a neutral LLM judge, Claude-opus-4-8, against the development ground truth on three integer dimensions, ranging from $0$ to $5$, that follow the official rubric. For MADU these dimensions are behavioural-semantic, spatial-topological, and temporal-structural. For MARE they are coarse-label, fine-label, and causal-reasoning. The judge sees only the question, the ground-truth answer, and the candidate answer, and is instructed to accept synonyms but penalise any reversal or omitted spatial or temporal detail. Each dimension is scaled to $[0,100]$ by a factor of $20$, and the development score is the mean of all twelve sub-scores. The open-ended ``score'' columns in \cref{tab:models,tab:ablate} report the mean of the resulting MADU and MARE dimensions on the $0$--$5$ scale.

\begin{table}[t]
\centering
\caption{Per-task model selection on the development split ($120$ questions). ``Closed'' is the mean closed-ended accuracy (struct-average, \%), and ``Open'' the mean MADU/MARE judge quality ($0$--$5$). \textbf{Bold} marks the best in each column.}
\label{tab:models}
\small
\begin{tabular}{lcc}
\toprule
Frozen model & Closed (acc.\ \%) $\uparrow$ & Open \\
\midrule
qwen3.7-plus    & 59.0          & \textbf{2.66} \\
qwen3.6-plus    & 51.0          & 2.60 \\
qwen3.5-plus    & 58.0          & 2.56 \\
kimi-k2.6       & 53.3          & 2.51 \\
gpt-5.5         & \textbf{61.0} & 2.62 \\
claude-opus-4.8 & 55.8          & 2.64 \\
\bottomrule
\end{tabular}
\end{table}

\begin{table}[t]
\centering
\caption{Prompt iterations on the development split. ``Closed dev.''\ is the struct-average closed-ended accuracy (\%) and ``Open dev.''\ the mean open-ended judge quality ($0$--$5$). $^{\dagger}$v5 is studied separately in \cref{sec:rcr-study}.}
\label{tab:prompts}
\small
\begin{tabularx}{\columnwidth}{l X c c}
\toprule
Version & Key change (targeted dimension) & Closed dev.\ $\uparrow$ & Open dev.\ $\uparrow$ \\
\midrule
v2 & fix output format (parseable answers) & 57.0 & 2.0 \\
v3 & add body-part, direction, and trend cues & 59.0 & 2.3 \\
v4 & rubric-aligned open-ended prompts (causal chain, near-neighbour exclusion, explicit labels) & 61.0 & 2.6 \\
v5 & recognition-conditioned reasoning$^{\dagger}$ & 61.0 & 2.7 \\
\bottomrule
\end{tabularx}
\end{table}

\begin{figure}[t]
\centering
\includegraphics[width=\linewidth]{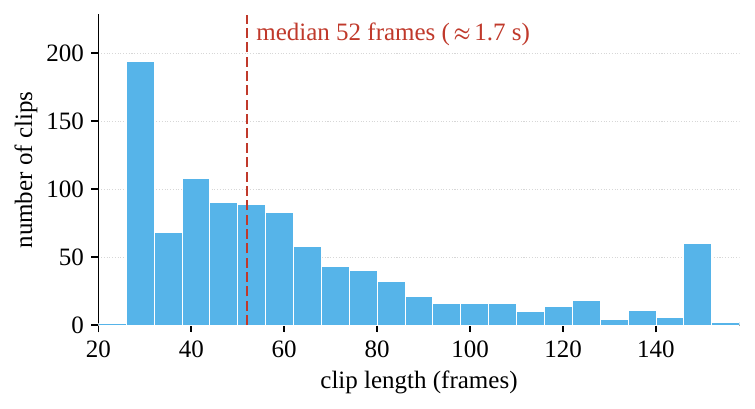}
\caption{Clip-length distribution of the $1{,}000$ MA-Bench videos (all $30$~fps). The median clip is $52$ frames. An $8$-frame sample covers only $\approx$ one in six frames of a typical clip.}
\label{fig:framedist}
\end{figure}

\begin{figure}[t]
\centering
\includegraphics[width=\linewidth]{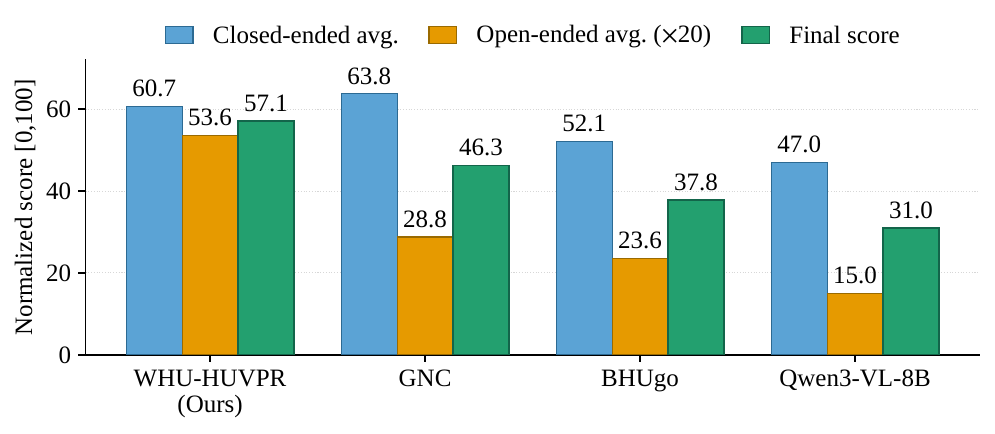}
\caption{Why our pipeline wins MA-Bench. The runner-up edges us on closed-ended recognition, but our open-ended description and reasoning are far stronger (open-ended average $\times 20$: $53.6$ vs.\ $28.8$), which decides the final score. Open-ended scores are scaled by $20$ for comparability.}
\label{fig:results}
\end{figure}

\subsection{Per-task model selection}
\label{sec:modelsel}
We benchmark six frozen MLLMs on the labelled development split and route each sub-task to the strongest, as \cref{tab:models} reports. On the closed-ended recognition sub-tasks gpt-5.5 is the most accurate, at a struct-average of $0.61$, whereas qwen3.7-plus produces the most faithful open-ended description and reasoning, at a mean judge quality of $2.66$ against $2.62$ for gpt-5.5. Our submission therefore routes the six closed-ended sub-tasks (CMAR, FMAR, SAD, MAD, MAS, PPR) to gpt-5.5 and the open-ended sub-tasks (MADU, MARE) to qwen3.7-plus. gpt-5.5 trails on the temporal-order sub-task MAS at $8$ frames but becomes the strongest once given a $64$-frame budget, as \cref{tab:ablate} shows. This per-task division is the decisive design choice behind our result. No single model is best everywhere, and the final-score gap to the runner-up is created entirely on the open-ended sub-tasks, which \cref{sec:track3main} analyses.

\subsection{Frame sampling and image fidelity}
\label{sec:frames}
The clips are short. All are recorded at $30$~fps with a median length of $52$ frames, about $1.7$~s, and a heavy tail to $152$ frames, as \cref{fig:framedist} shows. An $8$-frame sample covers only about one in six frames of a typical clip, so the right frame budget is \emph{task-dependent} rather than uniform. On gpt-5.5, supplying all sampled frames instead of $8$ has opposite effects across sub-tasks. It markedly helps the temporal sub-tasks, raising MAS from $0.58$ to $0.64$ and MAD from $0.54$ to $0.58$. It \emph{hurts} the static yes/no judgements, dropping SAD from $0.72$ to $0.62$ and PPR from $0.62$ to $0.48$. It leaves recognition (CMAR, FMAR) unchanged. The open-ended description and reasoning sub-tasks vary only slightly under either knob, and their best setting is again $8$ frames at quality $95$. \Cref{sec:rcr-study} corroborates the frame-count finding. We therefore allocate $64$ frames to the temporal sub-tasks MAS and MAD, and $8$ to the rest. Image \emph{fidelity} is a separate, orthogonal lever. Because micro-actions hinge on small, low-contrast cues, it is JPEG quality rather than frame count that preserves the spatial detail the fine-grained recognition sub-tasks depend on~\cite{zhang2025performing}. We therefore encode every frame at quality $95$, above the default $85$, at a modest payload cost. \Cref{tab:ablate} reports both sweeps. Its $8$-frame closed-ended rows use the quality-$95$ setting, which is immaterial for these non-recognition sub-tasks.

\begin{table}[t]
\centering
\caption{\textbf{Frame count and JPEG quality ablation} on the development split, organised by closed-ended and open-ended tasks. ``Rel.\ payload'' is the per-request image-token cost relative to the adopted $8$-frame quality-$95$ setting. \textbf{Bold} marks the chosen setting per task.}
\label{tab:ablate}
\small
\setlength{\tabcolsep}{4pt}
\begin{tabular}{lcccc}
\toprule
Sub-task  & Frames & JPEG Quality & Rel.\ Payload $\downarrow$ & Score $\uparrow$ \\
\midrule
\multicolumn{5}{c}{\cellcolor[HTML]{EFEFEF}{\emph{Close-End Tasks With Gpt-5.5} (accuracy $\uparrow$)}} \\
MAS & $8$ & $95$ & $1.0\times$ & $0.58$ \\
MAS & \textbf{64} & $95$ & $8.0\times$ & \textbf{0.64} \\
MAD & $8$ & $95$ & $1.0\times$ & $0.54$ \\
MAD & \textbf{64} & $95$ & $8.0\times$ & \textbf{0.58} \\
SAD & \textbf{8} & $95$ & $1.0\times$ & \textbf{0.72} \\
SAD & $64$ & $95$ & $8.0\times$ & $0.62$ \\
PPR & \textbf{8} & $95$ & $1.0\times$ & \textbf{0.62} \\
PPR & $64$ & $95$ & $8.0\times$ & $0.48$ \\
CMAR/FMAR & $8$ & $75$ & $0.43\times$ & $0.44$ \\
CMAR/FMAR & $8$ & $85$ & $0.54\times$ & $0.47$ \\
CMAR/FMAR & $8$ & \textbf{95} & $1.0\times$ & \textbf{0.48} \\
CMAR/FMAR & $32$ & $95$ & $4.0\times$ & $0.48$ \\
\midrule
\multicolumn{5}{c}{\cellcolor[HTML]{EFEFEF}{\emph{Open-End Tasks With Qwen3.7-plus} (judge $0$--$5$, $\uparrow$)}} \\
MADU/MARE  & $8$ & $75$ & $0.43\times$ & $2.58$ \\
MADU/MARE  & $8$ & $85$ & $0.54\times$ & $2.61$ \\
MADU/MARE  & \textbf{8} & \textbf{95} & $1.0\times$ & \textbf{2.66} \\
MADU/MARE  & $32$ & $95$ & $4.0\times$ & $2.64$ \\
MADU/MARE  & $64$ & $95$ & $8.0\times$ & $2.63$ \\
\bottomrule
\end{tabular}
\end{table}

\begin{table*}[t]
\centering
\caption{Official MAC~2026 Track~3 (MA-Bench) final leaderboard. Closed-ended columns are accuracy (\%). Open-ended columns (MADU/MARE, three dimensions each) are mean language-judge scores in $[0,5]$. The \textbf{Final} score averages all twelve sub-scores after normalising each to $[0,100]$ (open-ended $\times 20$). \textbf{Bold} marks the best value per column. Our pipeline wins despite a lower closed-ended average, driven by a large advantage on the open-ended description and reasoning tasks.}
\label{tab:track3}
\setlength{\tabcolsep}{4pt}
\begin{tabular}{l c | cccccc c | cccccc c}
\toprule
& & \multicolumn{7}{c|}{Closed-ended (Acc.\ \%)} & \multicolumn{7}{c}{Open-ended (judge score, $[0,5]$)} \\
Team & Final & CMAR & FMAR & SAD & MAD & MAS & PPR & CE-Avg & M.L1 & M.L2 & M.L3 & R.L1 & R.L2 & R.L3 & OE-Avg \\
\midrule
\textbf{WHU-HUVPR} & \textbf{57.14} & 61.00 & \textbf{48.70} & 67.75 & 59.70 & 64.70 & 62.50 & 60.73 & \textbf{2.36} & \textbf{2.83} & \textbf{2.47} & \textbf{3.10} & \textbf{2.63} & \textbf{2.67} & \textbf{2.68} \\
GNC & 46.27 & \textbf{62.50} & 45.00 & \textbf{76.70} & \textbf{68.00} & \textbf{67.15} & \textbf{63.30} & \textbf{63.78} & 1.47 & 1.55 & 1.46 & 1.56 & 1.38 & 1.23 & 1.44 \\
BHUgo & 37.83 & 40.00 & 34.20 & 63.35 & 54.30 & 62.25 & 58.60 & 52.12 & 0.97 & 1.03 & 0.86 & 1.64 & 1.34 & 1.21 & 1.18 \\
Qwen3-VL-8B & 30.99 & 32.23 & 37.11 & 56.10 & 53.69 & 47.94 & 54.74 & 46.97 & 0.50 & 0.51 & 0.47 & 1.18 & 1.00 & 0.84 & 0.75 \\
\bottomrule
\end{tabular}
\vspace{-0.5\baselineskip}
\end{table*}

\begin{figure*}[t]
\centering
\caption{Qualitative case study on a real MA-Bench clip. \emph{Top:} the eight uniformly sampled input frames. \emph{Bottom:} the actual questions with our pipeline's \emph{real}, abbreviated outputs, where the closed-ended sub-tasks are routed to gpt-5.5 for single-token answers and the open-ended MADU and MARE to qwen3.7-plus for structured description and reasoning.}
\vspace{0.5\baselineskip}
\includegraphics[width=\textwidth]{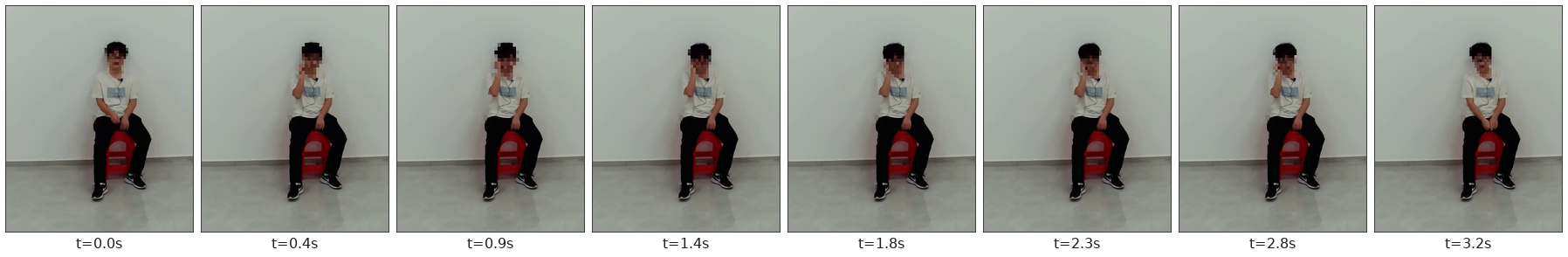}\\[5pt]
\footnotesize
\begin{tabularx}{\textwidth}{l l X l}
\toprule
Sub-task & Question (abbreviated) & Model output (real, abbreviated) & Model \\
\midrule
CMAR (coarse) & ``Which coarse-grained category? \texttt{A.}~Head\,\dots\,\texttt{C.}~Head-hand\,\dots\,\texttt{G.}~Upper limb'' & \texttt{C} (Head-hand) & gpt-5.5 \\
FMAR (fine) & ``Select the fine-grained category. \texttt{A.}~\dots\,touching face\,\dots\,\texttt{E.}~Touching ears'' & \texttt{E} (Touching ears) & gpt-5.5 \\
PPR (judgement) & ``In the early phase, does the \emph{left} hand noticeably approach the head?'' & \texttt{No} & gpt-5.5 \\
MADU (describe) & ``Describe the movements of each body part.'' & ``The right arm and hand raise in a single continuous arc to the side of the head: the elbow bends and the forearm travels diagonally up toward the right ear while the left hand stays static; a swift upward phase, a brief hold at the temple, then a downward return.'' & qwen3.7-plus \\
MARE (reason) & ``Give a step-by-step reasoning chain; identify the coarse- and fine-grained labels.'' & ``The right hand lifts in a smooth arc, contacts the ear/temple and holds; \emph{rules out} touching face (front of the face) and adjusting glasses (no contact) $\Rightarrow$ \textbf{coarse} Head-hand $\mid$ \textbf{fine} Touching ear.'' & qwen3.7-plus \\
\bottomrule
\end{tabularx}
\label{fig:case}
\end{figure*}

\subsection{Main Result: First Place on MA-Bench}
\label{sec:track3main}
\cref{tab:track3} reports the official final leaderboard. Our frozen Multimodal LLMs pipeline ranks \textbf{first} with a final score of $57.14$, ahead of the runner-up team by more than $10$ points. The decomposition is informative, on closed-ended recognition, we are competitive but \emph{behind} the runner-up team GNC ($60.73$ vs.\ $63.78$ average), yet on the open-ended description and reasoning tasks we lead every one of the six dimensions, lifting our open-ended average to $2.68$ against $1.44$ and $1.18$ for the other teams. \Cref{fig:results} visualises this gap. \Cref{fig:case} walks through a real clip, where the closed-ended route returns the labels Head-hand and Touching ear, and the open-ended route describes the spatial path and temporal phases and rules out the nearest look-alikes before reaching the same two labels. In other words, the win is carried by the quality and faithfulness of the \emph{generated} micro-action descriptions and explanations, exactly the capability that a strong frozen generative MLLM, prompted carefully and routed to the open-ended sub-tasks, is able to supply without any fine-tuning.

\subsection{Recognition-Conditioned Reasoning: A Gold-Grounded Ablation}
\label{sec:rcr-study}

We isolate the effect of recognition-conditioned reasoning, introduced in \cref{sec:rcr}, using the judge-free label-mention metric on the development split reasoning items. Specifically, we compare three prompting strategies for the generative MLLM: \emph{baseline} (no injected label), \emph{conditioned} (the discriminative MLLM's predicted coarse/fine label injected), and an \emph{oracle} upper bound (the label injected). Conditioning substantially improves reasoning performance, increasing coarse label accuracy from $28\%$ to $51\%$ and fine label accuracy from $7\%$ to $35\%$. When provided with the correct label, the oracle reaches $79\%$ and $66\%$ coarse/fine accuracy, respectively. These results demonstrate that the generative MLLM reliably incorporates the injected recognition signal into its reasoning process, while also revealing considerable headroom when recognition is perfect. Together, they indicate that the primary bottleneck lies in recognition accuracy rather than the generator's perceptual or reasoning capabilities. 
All experiments in this ablation are conducted on the development split.

\section{Conclusion}
We presented the training-free, first-place system for fine-grained micro-action understanding at the MAC~2026 challenge, built entirely on frozen multimodal LLMs. The pipeline routes each of the eight MA-Bench sub-tasks to the model best suited to it and wins by a wide margin on open-ended description and reasoning, using no fine-tuning and no ground-truth supervision. We further introduced recognition-conditioned reasoning, which decouples discrimination from explanation by conditioning a generator on an external discriminative prediction, and a judge-free, gold-grounded label-mention metric that sidesteps LLM-judge fluency bias. Together they localise the binding constraint on micro-action reasoning to label accuracy, which conditioning improves performance on development split data. 
More broadly, our results suggest that careful, training-free orchestration of frozen MLLMs is a strong and reproducible path for fine-grained micro-action understanding.
\begin{acks}
This work was supported by the NSFC Regional Innovation and Development Joint Fund under Grant U25A20537, and the National Key Research and Development Program of China under Grant 2024YFC3015600.
\end{acks}


\bibliographystyle{ACM-Reference-Format}
\bibliography{zotero_refs}


\end{document}